\documentclass[sigconf,10pt]{acmart}

\AtBeginDocument{%
  \providecommand\BibTeX{{%
    \normalfont B\kern-0.5em{\scshape i\kern-0.25em b}\kern-0.8em\TeX}}}

\acmBooktitle{October 2-6, 2023, Madrid, Spain} 
\acmPrice{15.00}
\acmISBN{978-1-4503-XXXX-X/07/08}
\copyrightyear{2023}
\acmYear{2023}
\setcopyright{rightsretained}
\acmConference[ACM MobiCom '23]{The 29th Annual International Conference on Mobile Computing and Networking}{October 2--6, 2023}{Madrid, Spain}
\acmBooktitle{The 29th Annual International Conference on Mobile Computing and Networking (ACM MobiCom '23), October 2--6, 2023, Madrid, Spain}\acmDOI{10.1145/3570361.3615741}
\acmISBN{978-1-4503-9990-6/23/10}

\usepackage{multirow}
\usepackage{tabularx}

\begin{document}

\title{Effective Abnormal Activity Detection on Multivariate Time Series Healthcare Data}

\author{Mengjia Niu}
\email{m.niu21@imperial.ac.uk}

\affiliation{%
  \institution{Imperial College London}
  \streetaddress{Exhibition Road}
  \city{London}
  \country{UK}
  \postcode{SW7 2AZ}
}

\author{Yuchen Zhao}
\email{yuchen.zhao@york.ac.uk}
\affiliation{%
  \institution{University of York}
  \streetaddress{Heslington}
  \city{York}
  \country{UK}
  \postcode{YO10 5GH}
}

\author{Hamed Haddadi}
\email{h.haddadi@imperial.ac.uk}
\affiliation{%
  \institution{Imperial College London}
  \streetaddress{84, WoodLane}
  \city{London}
  \country{UK}
  \postcode{W12 7SL}
}

\begin{abstract}
Multivariate time series (MTS) data collected from multiple sensors provide the potential for accurate abnormal activity detection in smart healthcare scenarios. However, anomalies exhibit diverse patterns and become unnoticeable in MTS data. Consequently, achieving accurate anomaly detection is challenging since we have to capture both temporal dependencies of time series and inter-relationships among variables. To address this problem, we propose a Residual-based Anomaly Detection approach, Rs-AD, for effective representation learning and abnormal activity detection. We evaluate our scheme on a real-world gait dataset and the experimental results demonstrate an $F_1$ score of 0.839.

\end{abstract}




\maketitle

\section{Introduction}

In the field of healthcare, identifying abnormal activities of individuals is important due to potentially unpredictable consequences. For instance, falls, particularly among the elderly, may result in injuries and, in some cases, even death if people do not get prompt treatment. With more and more sensors being deployed, there is growing potential to enhance the accuracy of anomaly detection services by leveraging multivariate time series (MTS) physiological data collected from these sensors. However, anomalies in MTS data are hard to identify since some implications of temporal dependency exist in each time series \cite{cook2019anomaly}. Moreover, abnormal patterns become unnoticeable in high-dimensional space and we have to identify inter-relationships among variables for detection tasks \cite{pang2021deep}. Thus, there is a need to develop methods which enable meaningful representation learning and effective modelling of both dependencies.

To address problems in processing MTS data, researchers have been exploring neural network-based methods. Recently, approaches based on deep neural networks (DNNs) have gained considerable attention due to their capacity for extracting data representations and performing downstream tasks including classification and anomaly detection \cite{wang2019deep}. However, most methods tend to focus solely on either abnormal temporal dependencies or adverse inter-relationships among various variables \cite{blazquez2021review}, leading to potential missing alerts for individuals. Recent advancements in modelling both types of anomalies \cite{zhang2019deep} have demonstrated the capability to fully leverage MTS data, showing superior performance compared to baseline methods. The researchers employed a multi-layered architecture comprising convolutional neural networks and attention-based convolutional long short-term memory (LSTM) networks, which are efficient in extracting high-level features but present a high expense of training time. 

\begin{figure}[t!]
    \centering
    \includegraphics[width=\linewidth]{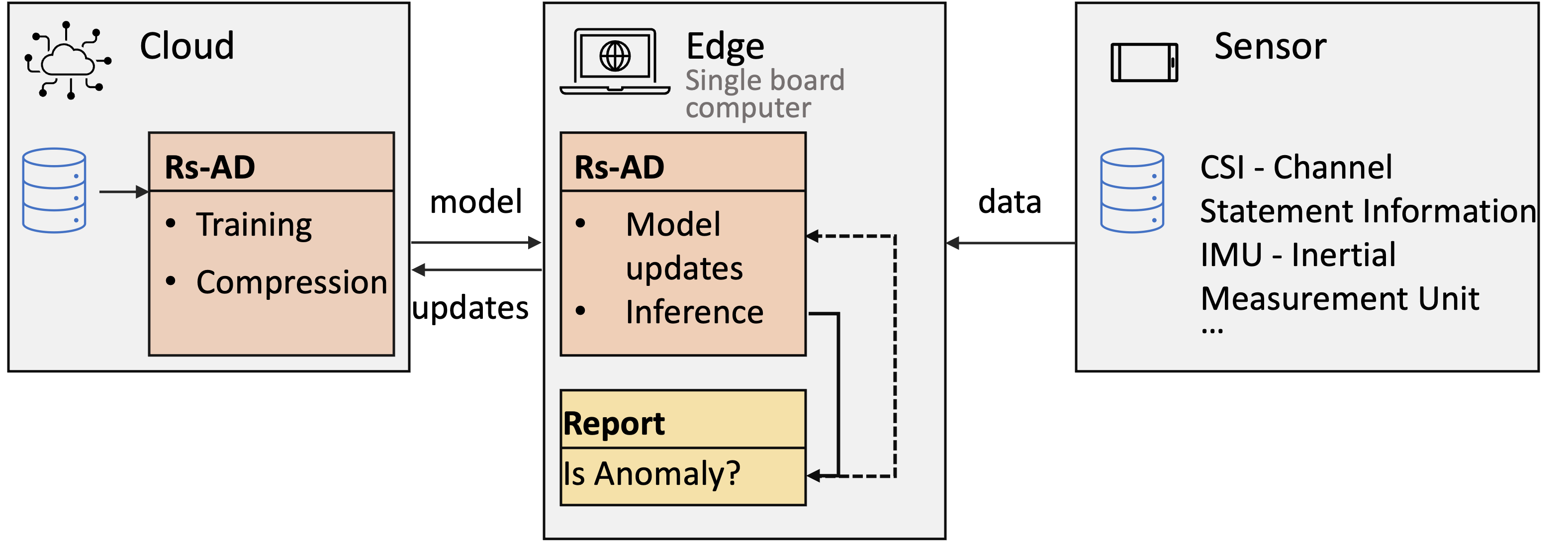}  
    \caption{Overview of an edge-intelligent system.}
    \label{FRAMEWORK}
\end{figure}

In this work, we propose a Residual-based Anomaly Detection (Rs-AD) model to detect intricate anomalies in MTS data. The proposed model is optimized using an objective function that incorporates multiple residuals, allowing for joint capture of temporal dependencies and inter-relationships among variables. Significantly, in our model, reconstruction data in the training process are also used as input data to facilitate meaningful representation learning. Therefore, our model achieves a smaller size compared with the model proposed in \cite{zhang2019deep} while preserving promising detection accuracy. We also present an extension of the deployment of the proposed approach in a general edge-intelligent system (as shown in Figure\ref{FRAMEWORK}), where the model will be trained and compressed in the cloud and subsequently sent to edge devices.

\section{Methodology}

\subsection{Problem Formulation}

\begin{figure}[t!]
    \centering
    \includegraphics[width=\linewidth]{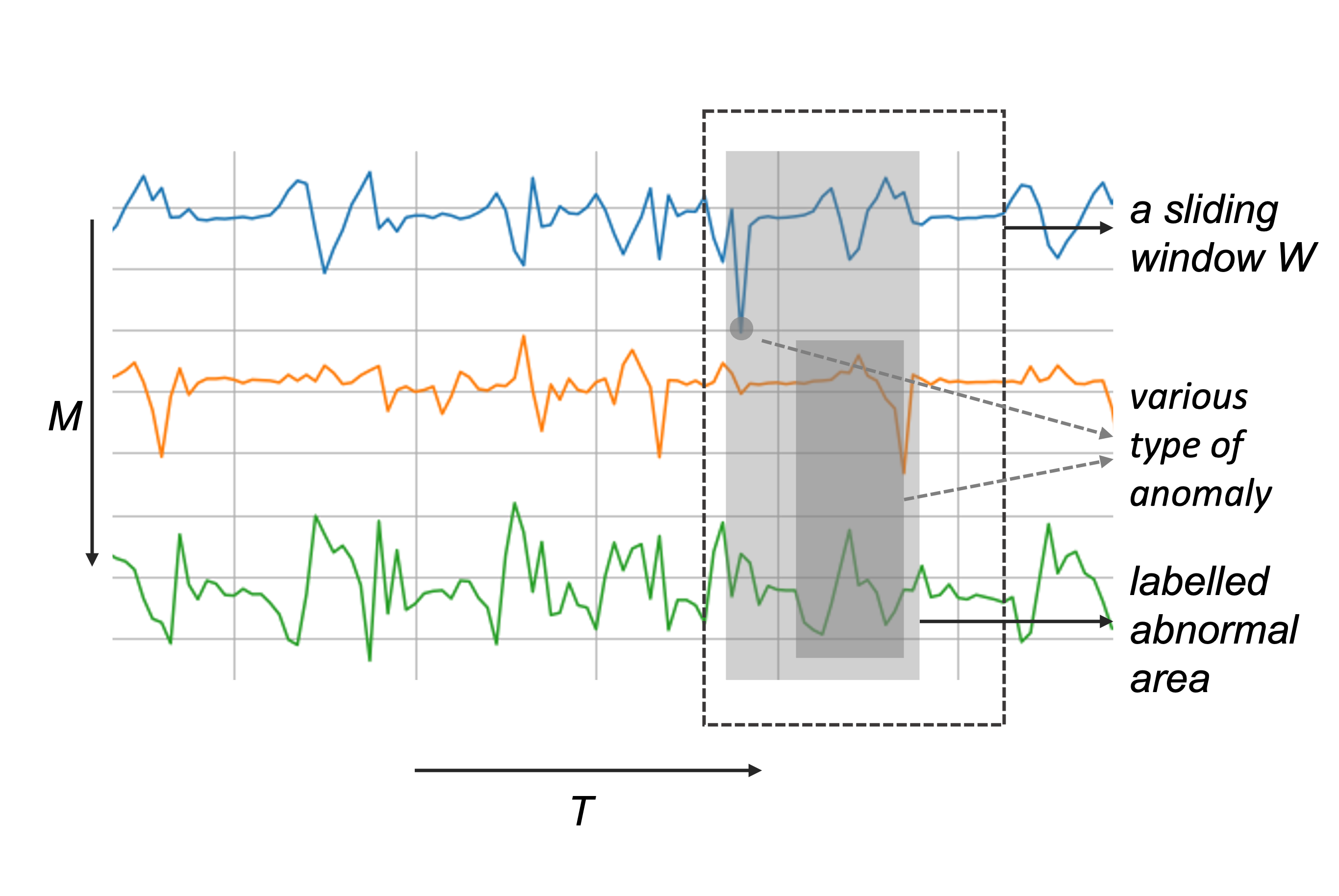}
    \caption{Formulation of MTS $x\in\mathbb{R}^{M\times T}$. A sliding window of length $W$ is given and the labelled grey area is an abnormal pattern. If a subsequence contains abnormal pieces, the whole subsequence will be defined as an anomaly.}
    \label{fig:MTS-definition}
\end{figure}

Data collected from various sensors can be concatenated and formulated as $\textbf{X} = (\textbf{x}_1,...,\textbf{x}_M)^T\in\mathbb{R}^{M\times T}$, where $M$ and $T$ are the numbers of variables and time stamps respectively \cite{li2021multivariate}. Given a sliding window of length $W$, the pattern of MTS can be denoted as 
\begin{math}
  \textbf{x}_\textbf{i} = (x_i^{(1)},x_i^{(2)},...,x_i^{(W)})
\end{math},
where $i\in\{1,2,...,M\}$. We aim to identify patterns that do not conform to normal patterns that make up the majority of data. Figure \ref{fig:MTS-definition} shows an example piece of MTS data and latent anomalies. 

\subsection{Framework Design}
\label{framework detail}

To process MTS data, the Rs-AD is proposed for anomaly detection. As discussed above, the key intuition is to identify abnormal patterns that may appear as abnormal temporal dependencies in single time series or inconsistent dependent relationships among various variables. Given these considerations, our approach aims to utilize reconstruction and multi-step prediction residuals to jointly learn two types of dependencies and identify abnormal patterns.

\begin{figure}[t!]
    \centering
    \includegraphics[width=\linewidth]{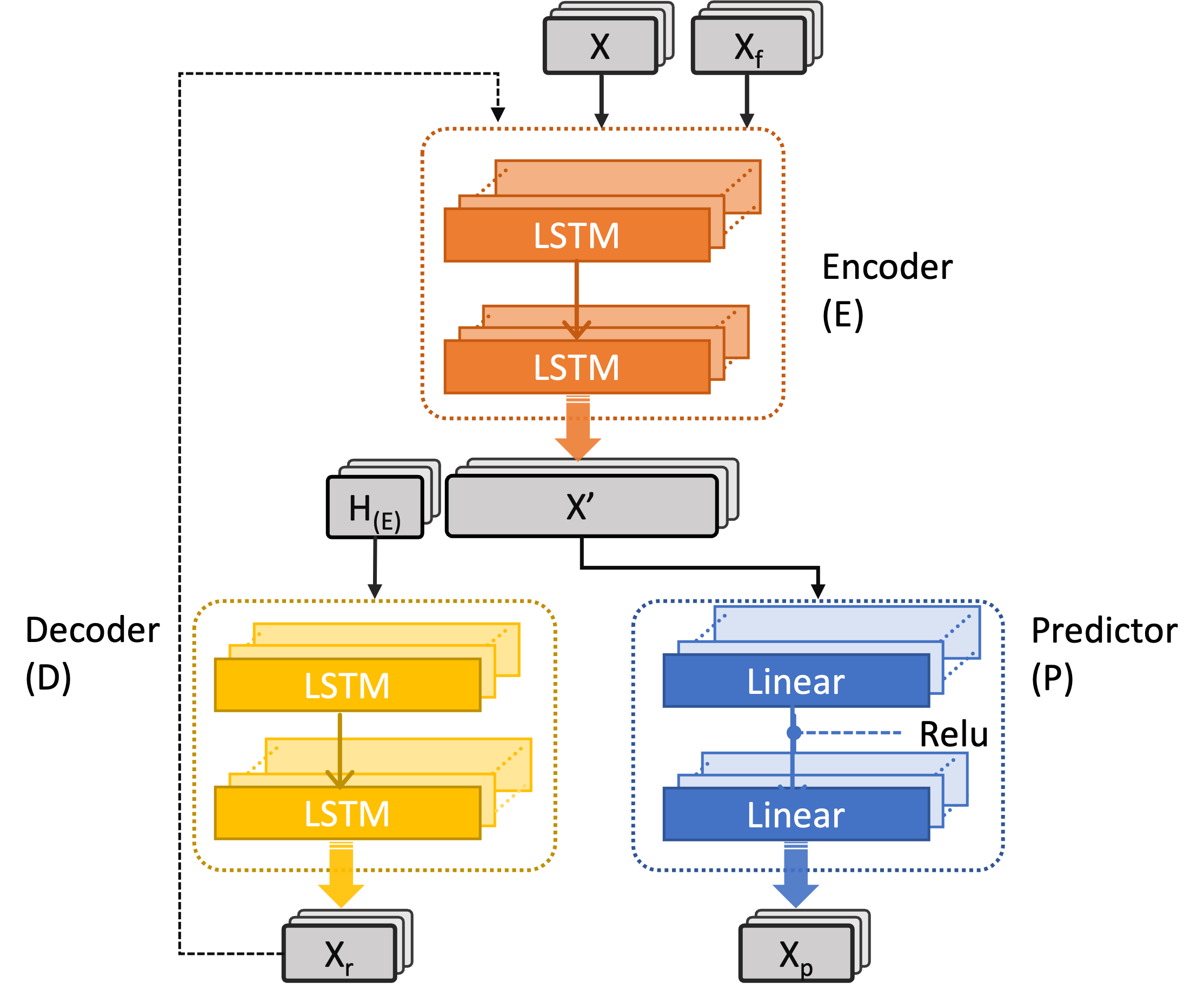}  
    \caption{Overview of the proposed detection model.}
    \label{model}
\end{figure}

As shown in Figure \ref{model}, Rs-AD consists of three main components. 
1) \textbf{Representation learning}, which employs an LSTM encoder (denoted as $E$) to capture both temporal and inter-relationship dependencies intuitively from input data $\textbf{X}$. 
2) \textbf{Data reconstruction} exploits an LSTM decoder (denoted as $D$) to reconstruct data $\textbf{X}_\textbf{r}$ from hidden states of the encoder. 
3) \textbf{Multi-step prediction} ($P$) learns to predict future data on the basis of a multilayer perceptron (MLP). The LSTM encoder and the MLP constitute the prediction network. The ground truth for multi-step prediction is denoted as $\textbf{X}_\textbf{r}$. Taking advantage of the framework, as in the research done by Audibert \textit{et al.} \cite{audibert2020usad}, inputs of the prediction network can be original data as well as reconstruction results of the decoder. The objective function for the whole model can be defined as: 
\begin{equation}
\begin{aligned}
\label{eq: Rs-AD loss}
&\mathcal{L}_{p1} = ||\textbf{X}_\textbf{f} - P(E(\textbf{X}))||_2, \\
&\mathcal{L}_{p2} = ||\textbf{X}_\textbf{f} - P(E(\textbf{X}_\textbf{r}))||_2,\\
&\mathcal{L} = \alpha * ||\textbf{X}-\textbf{X}_\textbf{r}||_2+ \beta * \mathcal{L}_{p1} + \gamma * \mathcal{L}_{p2},
\end{aligned}
\end{equation}

where $\alpha$, $\beta$ and $\gamma$ are hyperparameters used to fine-tune the relative influence of different residuals during the training process. Multiple residuals are concatenated into a score vector for each subsequence for anomaly measurements. Finally, we choose a threshold-based criterion to determine anomalies based on the score vector. 

Rs-AD will be extended to an edge-intelligent system as illustrated in Figure \ref{FRAMEWORK}. After being trained on the cloud, the model will be transmitted to an edge device, which is closer to data sources, to achieve precise and timely detection.

\section{Evaluation}
The evaluation of the effectiveness of the proposed approach to anomaly detection is as follows:

\textbf{Dataset} Daphnet dataset \cite{roggen2013uci} are used to evaluate the performance of the proposed method. This dataset was collected from ten patients with Parkinson's disease who were asked to wear three 3D-acceleration sensors and perform activities of daily living. During the experiment, these patients experienced abnormal gait freezing. We use data as suggested in \cite{schmidl2022anomaly} for gait freezing detection. Figure \ref{fig:MTS-definition} illustrates the data collected by a single sensor, depicting both normal activities and gait freezing. 

\textbf{Results} Considering the imbalance between normal and abnormal samples, we evaluate the performance via \textit{Recall (R)}, \textit{Precision (P)} and $F_1 score (F_1)$. Our model achieves an $F_1$ score of 0.839 and $R$ of 0.970, which demonstrate the promising detection performance of the proposed model. Additionally, we notice there is a compromise between $F_1$ and $R$.

\section{Conclusions and Future Work}

In this paper, we propose Rs-AD, a multiple-residual-based approach that enables modelling temporal dependencies and inter-relationships among variables for the accurate detection of diverse anomalies in smart healthcare environments. We evaluate the model using a real-world gait dataset, and the experimental results demonstrate its effectiveness in dealing with MTS data with an $F_1$ score of 0.839. In the future, we will deploy compressed Rs-AD on an edge-intelligent system, coupled with online model optimization techniques, to achieve personalized updates and enhance the detector's resilience to different monitored individuals. Furthermore, extensive experiments will be conducted on the edge to assess the reliability of inferences and evaluate any potential time latency.

\begin{acks}
Mengjia Niu is funded by Imperial College London and the China Scholarship Council (CSC). 
\end{acks}

\bibliographystyle{ACM-Reference-Format}
\bibliography{reference}

\end{document}